\documentclass{article}

\usepackage[preprint]{neurips_2025}

\usepackage[utf8]{inputenc} 
\usepackage[T1]{fontenc}    
\usepackage{hyperref}       
\usepackage{url}            
\usepackage{booktabs}       
\usepackage{amsfonts}       
\usepackage{nicefrac}       
\usepackage{microtype}      

\usepackage{algorithm}
\usepackage[noend]{algpseudocode}
\usepackage{amssymb}

\usepackage{graphicx}
\usepackage{subcaption}

\usepackage[table]{xcolor}

\usepackage{multirow}
\usepackage{tabularx}  

\usepackage{array}
\usepackage{booktabs}
\usepackage{makecell}
\usepackage{caption}

\renewcommand\arraystretch{1.3}

\usepackage{amsmath}

\usepackage{float}
\usepackage{graphicx}
\usepackage{svg} 

\usepackage{amsthm}
\theoremstyle{definition}

\usepackage[most]{tcolorbox}

\usepackage{multicol}
\usepackage{tcolorbox}
\usepackage{amsmath}
\usepackage{lipsum}

\tcbset{
  common/.style={colback=cyan!10, colframe=cyan!80!black, boxrule=0.5mm, sharp corners},
  diffA/.style={colback=green!10, colframe=green!80!black, boxrule=0.5mm, sharp corners},
  diffB/.style={colback=orange!10, colframe=orange!80!black, boxrule=0.5mm, sharp corners}
}

\newcommand{\coloredcell}[1]{%
  \ifdim #1 pt < 0.02 pt
    \cellcolor{red!10}#1
  \else
    \ifdim #1 pt < 0.025 pt
      \cellcolor{red!20}#1
    \else
      \ifdim #1 pt < 0.15 pt
        \cellcolor{red!40}#1
      \else
        \cellcolor{red!70}#1
      \fi
    \fi
  \fi
}

\newcommand{\methodnamefull}{Dynamic Adaptation of Reasoning Trajectories}
\newcommand{\methodname}{DART}

\title{Beyond Templates: Dynamic Adaptation of Reasoning Demonstrations via Feasibility-Aware Exploration}

\author{
\textbf{Yong Wu}$^{1}$\thanks{Equal contribution.} \quad
\textbf{Weihang Pan}$^{1}$\footnotemark[1] \quad
\textbf{Ke Li}$^{2}$ \quad
\textbf{Chen Binhui}$^{3}$ \quad
\textbf{Ping Li}$^{4}$ \quad
\textbf{Binbin Lin}$^{1}$\thanks{Corresponding author.} \\
$^{1}$College of Software Technology, Zhejiang University \\
$^{2}$Fullong Technology, Ningbo, Zhejiang, China \\
$^{3}$Ningbo Zhoushan Port Co., Ltd., Ningbo, Zhejiang, China \\
$^{4}$School of Computer Science and Technology, Hangzhou Dianzi University \\
\texttt{\{wu.yong, panweihang, binbinlin\}@zju.edu.cn}, \texttt{like@fullong.com.cn}, \\
\texttt{chenbinhui@126.com}, \texttt{lpcs@hdu.edu.cn}
}

\begin{document}

\maketitle

\begin{abstract}
    Large language models (LLMs) have shown remarkable reasoning capabilities, yet aligning such abilities to small language models (SLMs) remains a challenge due to distributional mismatches and limited model capacity. Existing reasoning datasets, typically designed for powerful LLMs, often lead to degraded performance when directly applied to weaker models. In this work, we introduce \methodnamefull{} (\methodname{}), a novel data adaptation framework that bridges the capability gap between expert reasoning trajectories and diverse SLMs. Instead of uniformly imitating expert steps, \methodname{} employs a \textit{selective imitation strategy} guided by step-wise adaptability estimation via solution simulation. When expert steps surpass the student's capacity---signaled by an \textit{imitation gap}---the student autonomously explores alternative reasoning paths, constrained by outcome consistency. We validate \methodname{} across multiple reasoning benchmarks and model scales, demonstrating that it significantly improves generalization and data efficiency over static fine-tuning. Our method enhances supervision quality by aligning training signals with the student’s reasoning capabilities, offering a scalable solution for reasoning alignment in resource-constrained models.
\end{abstract}


\section{Introduction}
\label{sec:introduction}


Large language models (LLMs) have recently achieved remarkable performance in complex reasoning tasks such as mathematics and programming(~\cite{openai2024o1systemcard, shao2024deepseekmathpushinglimitsmathematical}). A key insight from recent work(~\cite{zhou2024lima, yue2024mammoth2scalinginstructionsweb, ye2025limoreasoning}) is that small, high-quality instruction datasets are surprisingly effective at eliciting sophisticated reasoning abilities in large models. This discovery challenges traditional beliefs(~\cite{li2024numinamath, yu2024metamathbootstrapmathematicalquestions}) that complex cognitive skills necessarily require massive supervised fine-tuning, opening promising avenues for data-efficient model alignment.

Despite the remarkable effectiveness of small, high-quality instruction datasets in eliciting sophisticated reasoning, mainstream approaches(~\cite{zhou2024lima, ye2025limoreasoning,muennighoff2025s1simpletesttimescaling}) remain reliant on \textbf{static, pre-collected} reasoning datasets. While effective in controlled environments, these datasets struggle to generalize across heterogeneous pretraining distributions, particularly for small language models (SLMs) with diverse training data and limited reasoning capabilities~\cite{xu2024strongermodelsstrongerteachers, yeo2025demystifyinglongchainofthoughtreasoning}. Disparities in model scale, reasoning proficiency, and training history exacerbate distributional mismatches, significantly hindering the activation of reasoning skills.

To address these challenges, we introduce \textbf{\methodnamefull{} (\methodname{})}, a novel data adaptation framework designed to bridge the distribution gap between static reasoning datasets and diverse SLMs. Instead of enforcing uniform imitation of expert demonstrations, \methodname{} introduces a \textit{selective imitation} strategy guided by \textit{imitation feasibility estimate}. For each step provided by the expert, \methodname{} dynamically assesses the likelihood that the student model can successfully complete the reasoning process conditioned on adopting that step. When imitation is deemed infeasible, the student autonomously explores alternative trajectories while maintaining the consistency of the outcome with the objective of the original task. This approach enables \methodname{} to flexibly adapt high-quality reasoning datasets to heterogeneous model populations, significantly improving reasoning elicitation under distribution shift.

In summary, our contributions are as follows.
\begin{itemize}
    \item We identify the critical limitations of applying static curated reasoning datasets to diverse small language models and propose \textbf{\methodname{}}, a novel framework for adapted reasoning data guided by imitation feasibility.
    \item We introduce a Monte Carlo simulation-based method to estimate the feasibility of imitation per step, allowing selective supervision tailored to the student model capabilities.
    \item We develop an autonomous exploration mechanism that allows models to recover from infeasible supervision points, generating outcome-consistent alternative reasoning paths.
    \item Through extensive experiments across different model scales and benchmarks, we demonstrate that \methodname{} substantially improves reasoning performance over static fine-tuning, achieving superior data efficiency and generalization.
\end{itemize}


\section{Preliminaries and Limitations of Supervised Imitation on Expert Trajectories}
\label{sec:preliminaries}

\subsection{Problem Definition: Reasoning Capability Elicitation via Minimal Demonstrations}

We define the reasoning elicitation problem in the context of large language models (LLMs) with latent pre-trained knowledge. Let $\mathcal{Q}$ denote the space of reasoning problems, $\mathcal{A}$ the space of answers, and $\mathcal{R}$ the space of reasoning chains, where each $r \in \mathcal{R}$ is a sequence of logical steps $r = \{s_1, s_2, \ldots, s_n\}$.

The goal is to learn a reasoning function:
\begin{equation}
    f : \mathcal{Q} \rightarrow \mathcal{R} \times \mathcal{A}
    \label{eq:easoning_function:}
\end{equation}

so that, given a question $q \in \mathcal{Q}$, the model generates a logically valid reasoning chain $r \in \mathcal{R}$ and a verifiable final answer $a \in \mathcal{A}$.


Prior work (e.g., ~\cite{ye2025limoreasoning}, ~\cite{muennighoff2025s1simpletesttimescaling}) suggests that reasoning competence in large language models (LLMs) can be elicited not by scale alone, but through a small set of carefully crafted demonstrations that expose the underlying cognitive structure of reasoning. This paradigm assumes that latent reasoning skills embedded within pretrained models can be activated through appropriately designed prompts in the form of explicit multi-step exemplars.



Let $\mathcal{D} = \{(q_i, r_i, a_i)\}_{i=1}^N$ represent a compact yet high-quality dataset ($N \ll |\mathcal{Q}|$), where each tuple contains a question $q_i$, a structured reasoning chain $r_i$, and its corresponding answer $a_i$. Each $r_i$ serves as a \textbf{cognitive template}—an interpretable, step-wise reasoning demonstration designed to guide the model through logical steps with intermediate verification. Instead of introducing new knowledge, these templates activate the model's latent reasoning capabilities by leveraging structured prompting~\cite{wei2022chain,zhou2024lima,ye2025limoreasoning}.

\subsection{Limitations of Supervised Imitation on Expert Demonstrations}

Despite its pedagogical appeal, supervised imitation over expert demonstrations exhibits critical limitations when applied to LLMs with diverse capacity levels.


This paradigm(~\cite{wei2022chain,ye2025limoreasoning}) assumes that the model possesses sufficient latent competence to internalize and reproduce the reasoning trajectory in each template. In practice, this assumption frequently fails. A template $r_i$ may (i) over-challenge the model by invoking reasoning procedures not encoded in its weights, or (ii) misalign with the model’s inductive biases, causing representational mismatch. We define a reasoning failure event $\mathcal{F}$ as the inability of the model to emulate the intended behavior given an input-template pair:
\begin{equation}
    \mathcal{F}(f; q, r, a) = \mathbb{I}[f(q) \not\approx (r, a)]
    \label{eq:reasoning_failure}
\end{equation}
where $\mathbb{I}[\cdot]$ is the indicator function. Such failures may arise from superficial imitation, incomplete reasoning chains, or insufficient justification for the final answer. 

Compounding this challenge is the substantial cost associated with constructing template datasets $\mathcal{D}$ that satisfy the Cognitive Template Demonstration criterion. Such templates demand meticulous logical decomposition, intermediate verification, and fine-grained pedagogical design. Furthermore, a template crafted for a specific model often fails to generalize to others due to differences in scale, pretraining corpus, or architectural inductive biases, resulting in pronounced distributional shifts. As highlighted in prior work on imitation learning(~\cite{Pomerleau1991,RossBagnell2010}), relying on static datasets for training can lead to a distribution mismatch between the output sequences encountered during training and those generated auto-regressively by the student at inference time, undermining generalization and robustness.

\paragraph{The Need for Imitation Feasibility-Aware Adaptation.}  
These limitations highlight the inadequacy of static demonstrations in addressing the diversity of model behaviors. We argue for a dynamic grounding mechanism that aligns template presentation with the target model’s internal capacity and abstraction level. Rather than treating $\mathcal{D}$ as fixed input, the elicitation process should adaptively align the demonstrated reasoning path with the model's own preferred or accessible inference trajectories, potentially reformulating how the reasoning unfolds to match internal representations. This motivates our central question:

\begin{quote}
\textit{Can we design a dynamic adaptation mechanism that reliably anchors cognitive templates in model-specific latent space, enabling scalable and robust reasoning?}
\end{quote}

In the following section, we instantiate this motivation via our proposed framework — \textbf{\methodnamefull{} (\methodname{})}.


\section{Methodology}
\label{sec:methodology}
\begin{figure}[t]
    \centering                                    
    \includegraphics[width=\linewidth]{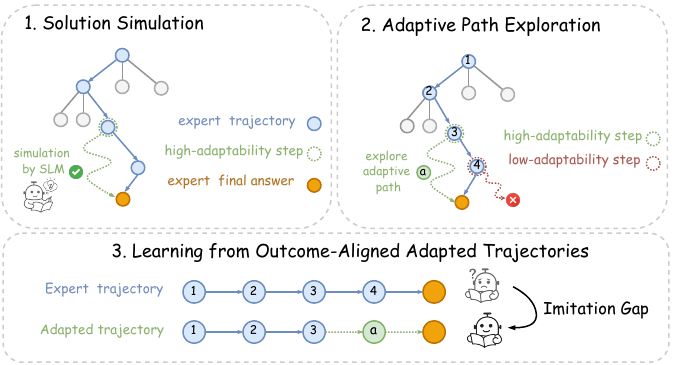}    
    \caption{Overview of the \methodname{} framework.} 
    \label{fig:method_pipeline}
\end{figure}

In this section, we propose \textbf{\methodnamefull{} (\methodname{})}, a capability-aware adaptation framework designed to align expert-level reasoning data with the capacity of small language models (SLMs). Instead of statically mimicking expert trajectories from the elicitation template set, \methodname{} introduces a selective imitation mechanism that dynamically adapts supervision signals based on the model's reasoning proficiency. The framework comprises three key components: (1) step-wise adaptability estimation via solution simulation (Section~\ref{sec:mcts_feasibility}), (2) imitation gap detection and adaptive path exploration (Section~\ref{sec:adaptive_exploration}), and (3) learning from outcome-aligned adapted trajectories (Section~\ref{sec:distillation}). Figure~\ref{fig:method_pipeline} provides an overview of the pipeline.


\subsection{Step-wise Adaptability Estimation via Solution Simulation}
\label{sec:mcts_feasibility}

To determine whether a given expert step is suitable for imitation, we introduce the concept of \textbf{adaptability}: the likelihood that a student model can reach the correct answer when conditioned on that step. This evaluation is conducted via solution simulation—akin to Monte Carlo Tree Search (~\cite{kocsis2006bandit,silver2016mastering,swiechowski2023monte})—by rolling out multiple completions from partially constructed trajectories that incorporate the candidate step.

Let \( s_{<t} = \{s_0, s_1, \ldots, s_{t-1}\} \) be the prefix of expert steps, and \( s_t \) the candidate step under evaluation. The adaptability score \( f_t \) is computed as:

\begin{equation}
    f_t = Q(s_{<t}, s_t) = \frac{1}{N_{\text{sim}}} \sum_{i=1}^{N_{\text{sim}}} \mathbb{I}(a_i^{\text{final}} = a^*)
    \label{eq:adaptability}
\end{equation}

where \( N_{\text{sim}} \) denotes the total number of \textit{rollouts} performed for each candidate step \( s_t \), with each rollout simulating a complete reasoning trajectory conditioned on the prefix \( s_{<t} \) and the adoption of step \( s_t \).

Empirically observed patterns (see Section~\ref{sec:imitation_gap}) suggest that adaptability tends to rise in the early stages of expert trajectories, but drops sharply beyond a certain point. This non-monotonic behavior motivates our definition of the \textbf{imitation gap}, a regime in which continued imitation becomes counterproductive due to the increasing complexity of the remaining expert steps.

\subsection{Adaptive Path Exploration}
\label{sec:adaptive_exploration}
To avoid overfitting to brittle expert demonstrations, we monitor the \textit{adaptability score} throughout the trajectory and halt imitation once a significant drop is detected (see Equation~\ref{eq:adaptability}). Motivated by the need to overcome low-adaptability segments that may hinder generalization, \methodname{} transitions to autonomous rollout beyond the gap, generating a continuation from the last high-adaptability prefix:
\begin{equation}
    \tau_{\text{adapt}} = (s_0, s_1, \ldots, s_{t-1}, s'_{t}, s'_{t+1}, \ldots, s'_T),
\end{equation}
where \( s'_t, \ldots, s'_T \) are student-generated reasoning steps. Inspired by outcome-based learning strategies(~\cite{deepseekai2025r1}), we do not constrain this trajectory to mimic the expert's form. Instead, we enforce an \emph{outcome consistency} constraint to ensure semantic alignment,as described in Eq.~\eqref{eq:outcome},as we observe that process supervision(~\cite{lightman2024lets,zhang2025processreward},), such as via a Process Reward Model (PRM), often encounters inherent ambiguities and standardization challenges in practice.
\begin{equation}
    C(\tau_{\text{adapt}}, \tau_{\text{expert}}) = 
    \begin{cases} 
    1, & \text{if } \mathcal{O}(\tau_{\text{adapt}}) = \mathcal{O}(\tau_{\text{expert}}), \\
    0, & \text{otherwise}.
    \end{cases}
    \label{eq:outcome}
\end{equation}
Here, \( C \in \{0,1\} \) denotes task-level agreement, with \( \mathcal{O}(\cdot) \) representing the final answer obtained by executing a reasoning path. Specifically, \( \mathcal{O}(\tau_{\text{expert}}) \) refers to the outcome of the expert demonstration, while \( \mathcal{O}(\tau_{\text{adapt}}) \) captures the result of the student’s adapted trajectory. The constraint \( \mathcal{O}(\tau_{\text{adapt}}) = \mathcal{O}(\tau_{\text{expert}}) \) ensures that, although the reasoning paths may differ, their semantic outcomes are equivalent. This outcome consistency criterion allows the student to depart from brittle expert traces while preserving task correctness.

This strategy empowers the student model to develop its own reasoning strategies beyond segments with low adaptability, guided solely by the correctness of the final outcome. By anchoring supervision at the outcome level rather than mimicking intermediate steps, we alleviate the brittleness of process-level imitation. This encourages robust generalization, reduces reliance on ambiguous or inconsistent expert demonstrations, and aligns with the broader goal of enabling flexible yet goal-directed reasoning.

\begin{algorithm}[t]
\caption{\methodname{}: \methodnamefull{}}
\label{alg:adaptive_trajectory}
\begin{algorithmic}[1]
\Require Expert trajectory $\tau_{\text{expert}} = \{s_0, s_1, \ldots, s_T\}$, student model $\pi_{\text{student}}$, ground-truth answer $a^*$, adaptation simulation count $N_{\text{sim}}$
\State Initialize prefix $\tau_{\text{prefix}} \gets \emptyset$; adaptability scores $\mathcal{F} \gets []$
\For{$t = 0$ to $T$}
    \State Compute adaptability score $f_t \gets Q(s_{<t}, s_t)$ \Comment{See Eq.~\eqref{eq:adaptability}}
    \State Append $f_t$ to $\mathcal{F}$
\EndFor
\State Find $t_{\text{peak}}$ where $f_t$ attains its local maximum;
\[
t_{\text{gap}} \gets \min\left\{ t > t_{\text{peak}} \mid f_t < f_{t_{\text{peak}}} - \epsilon \right\}
\quad \text{where } \epsilon > 0 \text{ defines a significant drop threshold}
\]
\State Truncate expert prefix: $\tau_{\text{prefix}} \gets \{s_0, \ldots, s_{t_{\text{gap}} - 1}\}$
\State Initialize adapted trajectory: $\tau_{\text{adapt}} \gets \tau_{\text{prefix}}$

\While{not terminal}
    \State Sample next step $s' \sim \pi_{\text{student}}(\cdot \mid \tau_{\text{adapt}})$
    \State Append $s'$ to $\tau_{\text{adapt}}$
\EndWhile

\If{$\mathcal{O}_{\text{adapt}}(\tau_{\text{adapt}}) = \mathcal{O}_{\text{expert}}(\tau_{\text{expert}})$}
    \State Retain $\tau_{\text{adapt}}$ for distillation \Comment{See Eq.~\eqref{eq:distillation}}
\Else
    \State Discard $\tau_{\text{adapt}}$
\EndIf
\end{algorithmic}
\end{algorithm}

\subsection{Learning from Outcome-Aligned Adapted Trajectories}
\label{sec:distillation}


To effectively activate the student model's own reasoning ability, we apply a standard cross-entropy loss on the outcome-aligned adapted trajectories generated during autonomous exploration. This training objective encourages the model to reinforce reasoning patterns that are not only aligned with the task goal but also feasible under its own capacity.

Training proceeds by distilling the adapted trajectory $\tau_{\text{adapt}}$ using a standard cross-entropy loss(~\cite{kimrush2016sequence,Bengio2003neuralprobabilistic}):
\begin{equation}
    L_{\text{\methodname{}}} = - \sum_{t=1}^{T} \mathbb{E}_{(s_{<t}, a_t) \sim \tau_{\text{adapt}}} \left[ \log \pi_{\text{student}}(a_t \mid s_{<t}) \right]
    \label{eq:distillation}
\end{equation}
Here, $s_{<t} = \{s_0, \ldots, s_{t-1}\}$ denotes the contextual prefix consisting of all prior reasoning steps up to time $t$, and $a_t$ is the corresponding next-step decision. This loss encourages the student model $\pi_{\text{student}}$ to maximize the likelihood of producing $a_t$ when conditioned on its own reasoning history.

By learning from outcome-aligned yet model-compatible trajectories, \methodname{} provides high-quality supervision that reflects the student’s actual competence. This approach decouples the training signal from rigid trajectory matching, improving both robustness and scalability across models with varying capacity.

\begin{table}[htbp]
  \centering
  \normalsize
  \caption{Accuracy (\%) on LIMO and Math-QwQ-32B across adaptation strategies and model sizes. \textbf{Static} overfits to noisy data, while \textbf{Adaptation-Full} improves results through exploration and filtering of low-adaptability segments. \textbf{Bold} denotes the best, and \underline{underline} the second-best.}
  \resizebox{\textwidth}{!}{%
  \begin{tabular}{llccccccccc}
    \toprule
    \thead{Model} & \thead{Method} & \thead{Data Size} & \thead{GSM8K} & \thead{MATH} & \thead{Minerva\\Math} & \thead{GaoKao\\2023 En} & \thead{Olympiad\\Bench} & \thead{College\\Math} & \thead{MMLU\\STEM} & \thead{Avg.} \\
    \midrule
    \multicolumn{11}{c}{\underline{\textbf{LIMO Dataset (817 samples)}}}\\
    \multirow{3}{*}{0.5B}
      & No-Tuning             & 817 & 49.1 & \underline{34.2} & \underline{6.6} & \underline{30.4} & \underline{9.3} & \underline{28.9} & \underline{36.7} & \underline{27.9}\\
      & Static                & 817 & \underline{49.6} & 32.9 & 5.1 & 26.8 & 7.7 & 27.3 & 32.9 & 26.0\\
      & \textbf{Adaptation-Full} &  \textbf{202}& \textbf{52.2} & \textbf{35.2} & \textbf{8.5} & \textbf{32.5} & \textbf{9.8} & \textbf{29.1} & \textbf{37.2} & \textbf{29.2}\\
    \midrule
    \multirow{3}{*}{1.5B}
      & No-Tuning             & 817 & 73.5 & \textbf{55.4} & \underline{15.8} & \underline{48.6} & \textbf{21.6} & 38.5 & \underline{57.7} & \underline{44.4}\\
      & Static                & 817 & \textbf{75.2} & 54.4 & 15.8 & 48.6 & 19.6 & \underline{39.4} & 56.0 & 44.1\\
      & \textbf{Adaptation-Full} &  \textbf{546}& \underline{75.1} & \underline{55.2} & \textbf{16.9} & \textbf{49.6} & \underline{20.6} & \textbf{40.1} & \textbf{58.3} & \textbf{45.1}\\
    \midrule
    \multirow{3}{*}{3B}
      & No-Tuning             & 817 & \underline{66.6} & \underline{27.6} & \underline{56.6} & \underline{27.3} & 39.9 & 47.6 & \underline{27.3} & \underline{50.4}\\
      & Static                & 817 & 59.8 & 24.6 & 53.8 & 25.2 & \underline{41.8} & \underline{54.8} & 25.2 & 49.3\\
      & \textbf{Adaptation-Full} &  \textbf{621}& \textbf{66.6} & \textbf{28.3} & \textbf{59.5} & \textbf{30.4} & \textbf{43.9} & \textbf{62.7} & \textbf{30.4} & \textbf{54.1}\\
    \midrule
    \multicolumn{11}{c}{\underline{\textbf{Math-QwQ-32B Dataset (5383 samples)}}}\\
    \multirow{3}{*}{0.5B}
      & No-Tuning             & 5383& \underline{49.1} & \underline{34.2} & \underline{6.6} & \underline{30.4} & \underline{9.3} & \textbf{28.9} & \underline{36.7} & \underline{27.9}\\
      & Static                & 5383& 39.8 & 20.5 & 2.6 & 20.5 & 5.9 & 17.3 & 27.9 & 19.2\\
      & \textbf{Adaptation-Full} & \textbf{1829}& \textbf{49.6} & \textbf{35.2} & \textbf{7.0} & \textbf{30.9} & \textbf{9.3} & \underline{27.5} & \textbf{37.5} & \textbf{28.1}\\
    \midrule
    \multirow{3}{*}{1.5B}
      & No-Tuning             & 5383& \underline{73.5}& \textbf{55.4} & 15.8 & \underline{48.6}& \textbf{21.6} & \underline{38.5}& \underline{57.7}& \underline{44.4}\\
      & Static                & 5383& 64.7 & 38.8 & 13.6 & 34.8 & 12.3 & 28.5 & 43.4 & 33.7\\
      & \textbf{Adaptation-Full} & \textbf{3922}& \textbf{74.2} & \underline{55.1} & \textbf{17.6} & \textbf{48.6} & \underline{19.6} & \textbf{39.4} & \textbf{57.7} & \textbf{44.6} 
\\
    \midrule
    \multirow{3}{*}{3B}
      & No-Tuning             & 5383& \textbf{87.0}& \textbf{66.6} & \textbf{27.6} & \underline{56.6} & \underline{27.3} & \underline{39.9} & 47.6 & \underline{50.4}\\
      & Static                & 5383& 82.0 & 49.7 & 21.0 & 46.2 & 20.1 & 35.7 & \textbf{51.3} & 43.7\\
      & \textbf{Adaptation-Full} & \textbf{4100}& \underline{86.6} & \underline{65.3} & \underline{26.5} & \textbf{57.9} & \textbf{29.0} & \textbf{44.4} & \underline{50.6} & \textbf{51.5}\\
    \bottomrule
  \end{tabular}
  }
  \label{tab:elegant_adaptation_results}
  \vspace{-0.6cm} 
\end{table}

\section{Experiments}
\label{sec:experiments}

    

We evaluate \methodname{} across a series of mathematical reasoning benchmarks to assess its effectiveness in adapting expert data to student models of varying capacities. 

\subsection{Experimental Setup}


\paragraph{Adaptation Datasets.} 
We conduct adaptation experiments using two datasets. (1) \textbf{LIMO} dataset(~\cite{ye2025limoreasoning}), a curated set of 817 high-quality math reasoning examples with multi-step CoT demonstrations tailored.We use the official filtered release\footnote{\url{https://huggingface.co/GAIR/LIMO}}. 
(2) The \textbf{Math-QwQ-32B} dataset is a synthetic dataset derived from the MATH benchmark~\cite{MATHBENCH}, in which the Qwen/QwQ-32B-Preview model\footnote{\url{https://huggingface.co/Qwen/QwQ-32B-Preview}} generates long-form Chain-of-Thought (CoT) solutions for 5,383 problems from the training subset.

\paragraph{Adaptation Strategies.} We evaluate three adaptation strategies to disentangle the effects of selective imitation and adaptive exploration. (1) \textit{No-Tuning} denotes direct zero-shot evaluation. (2) \textit{Static} reflects standard offline supervised fine-tuning on the full set of expert trajectories, without any filtering or adaptability mechanism. (3) The \textit{Adaptation-Full} strategy represents the complete \methodname{} pipeline, integrating imitation gap detection with outcome-consistent student exploration. This approach empowers the model to autonomously explore alternative reasoning paths when expert imitation becomes unreliable. If the model can’t find a suitable alternative path, it discards that expert example. We evaluate our method on Qwen2.5-Instruct models at 0.5B\footnote{\url{https://huggingface.co/Qwen/Qwen2.5-0.5B-Instruct}}, 1.5B\footnote{\url{https://huggingface.co/Qwen/Qwen2.5-1.5B-Instruct}}, and 3B\footnote{\url{https://huggingface.co/Qwen/Qwen2.5-3B-Instruct}} scales, covering a diverse range of SLMs. All experiments are conducted using
NVIDIA A100 GPUs


\vspace{-0.3cm} 
\paragraph{Benchmark Tasks.} 
We evaluate \methodname{} on seven diverse benchmarks encompassing a broad spectrum of mathematical reasoning. These include GSM8K~\cite{gsm8k} and MATH~\cite{MATHBENCH}, covering grade-school to competition-level problems, and Minerva Math~\cite{lewkowycz2022solving}, which targets advanced symbolic reasoning. To assess linguistic and cultural generalization, we incorporate GaoKao 2023 En~\cite{mario}, a Chinese national exam benchmark. OlympiadBench~\cite{he-etal-2024-olympiadbench} features high-difficulty, compositional problems from international math competitions. College Math~\cite{tang2024mathscale} probes undergraduate-level topics in calculus, algebra, and discrete math. MMLU-STEM~\cite{hendrycks2021measuring} evaluates STEM-focused reasoning breadth. Overall adaptation is quantified by the arithmetic mean (Avg.) across all benchmarks.


\subsection{Main Results}

Table~\ref{tab:elegant_adaptation_results} reports performance across seven mathematical reasoning benchmarks, demonstrating the effectiveness of the proposed \methodname{} framework in aligning expert reasoning with the capabilities of small language models (SLMs).




\paragraph{Static Results} Static serves as a baseline that rigidly replicates expert trajectories without adaptation. Despite leveraging expert demonstrations, it underperforms even relative to models without any tuning, due to its inability to adjust to the limited capacity of small models. On \textit{Math-QwQ-32B} dataset, \textit{Static} underperforms by 8.7, 10.7, and 6.7 points at the 0.5B, 1.5B, and 3B scales, respectively.


\paragraph{Adaptation-Full Results} We evaluate \textbf{Adaptation-Full}, which enables active exploration when encountering imitation gaps, allowing models to deviate from expert trajectories. As shown in Table\ref{tab:elegant_adaptation_results}, it consistently outperforms static approaches across benchmarks. On the \textit{LIMO} dataset, it achieves gains of +3.2\%, +1.0\%, and +4.7\%, while on the \textit{Math-QwQ-32B} dataset, it improves by +8.5\%, +6.7\%, and +7.8\% at 0.5B, 1.5B, and 3B scales, respectively. These results highlight its effectiveness in aligning reasoning paths with model capacity.

\section{Analysis}
\label{sec:analysis}
We further analyze the internal mechanisms of \methodname{}, aiming to understand why selective imitation and autonomous exploration improve reasoning capabilities.

\subsection{Step-wise Adaptability Reveals the Emergence of the Imitation Gap}
\label{sec:imitation_gap}

\begin{figure}[t]
    \centering
    \caption{
    Step-wise adaptability scores across expert trajectories for student models of varying sizes (0.5B, 1.5B, 3B parameters) under LIMO (top row) and Math-QwQ-32B dataset (bottom row) supervision. The emergence of the \textbf{Imitation Gap} is evident: initial steps yield positive adaptation, but continued step-by-step imitation can become harmful.
    }
    \captionsetup[sub]{font=scriptsize}
    \begin{subfigure}[b]{0.32\linewidth}
        \centering
        \includegraphics[width=\linewidth]{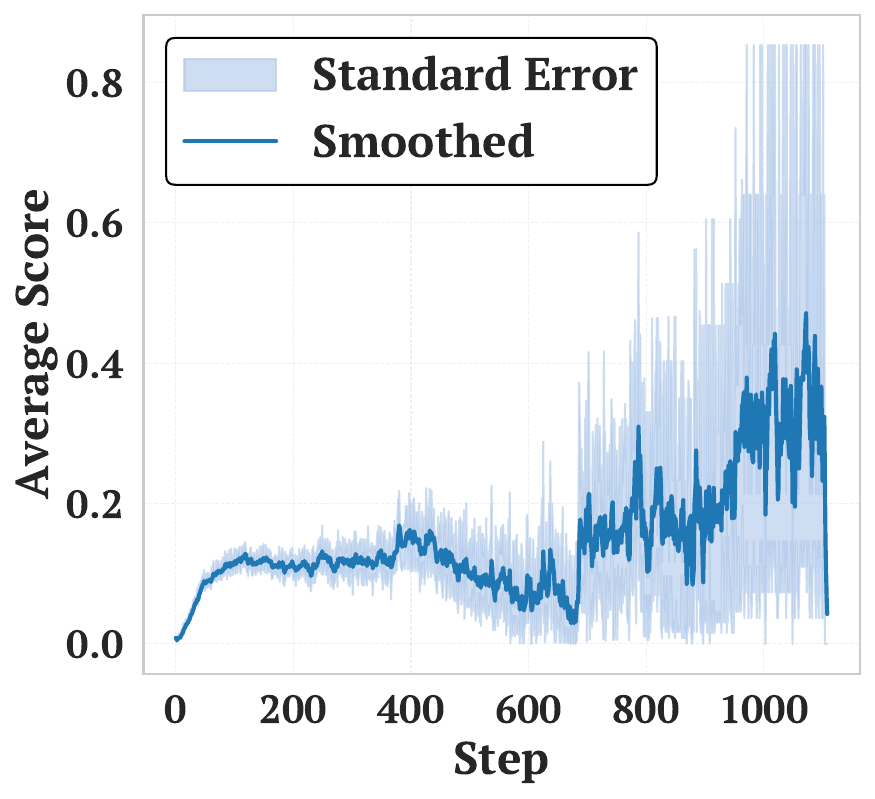}
        \caption{Qwen2.5-0.5B-Instruct (LIMO)}
    \end{subfigure}
    \begin{subfigure}[b]{0.32\linewidth}
        \centering
        \includegraphics[width=\linewidth]{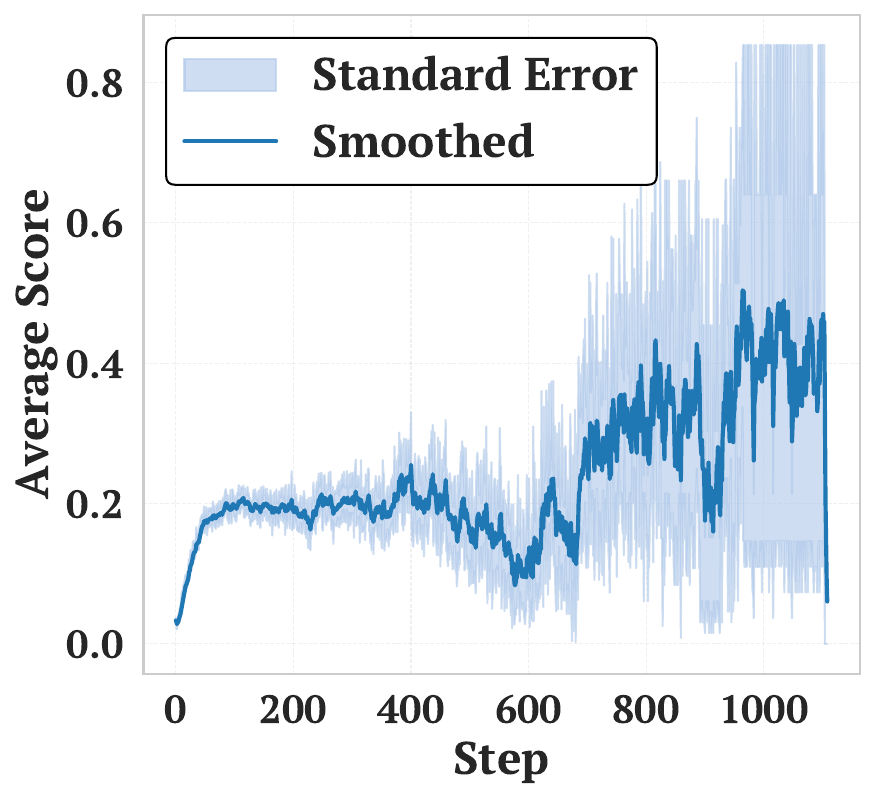}
        \caption{Qwen2.5-1.5B-Instruct (LIMO)}
    \end{subfigure}
    \begin{subfigure}[b]{0.32\linewidth}
        \centering
        \includegraphics[width=\linewidth]{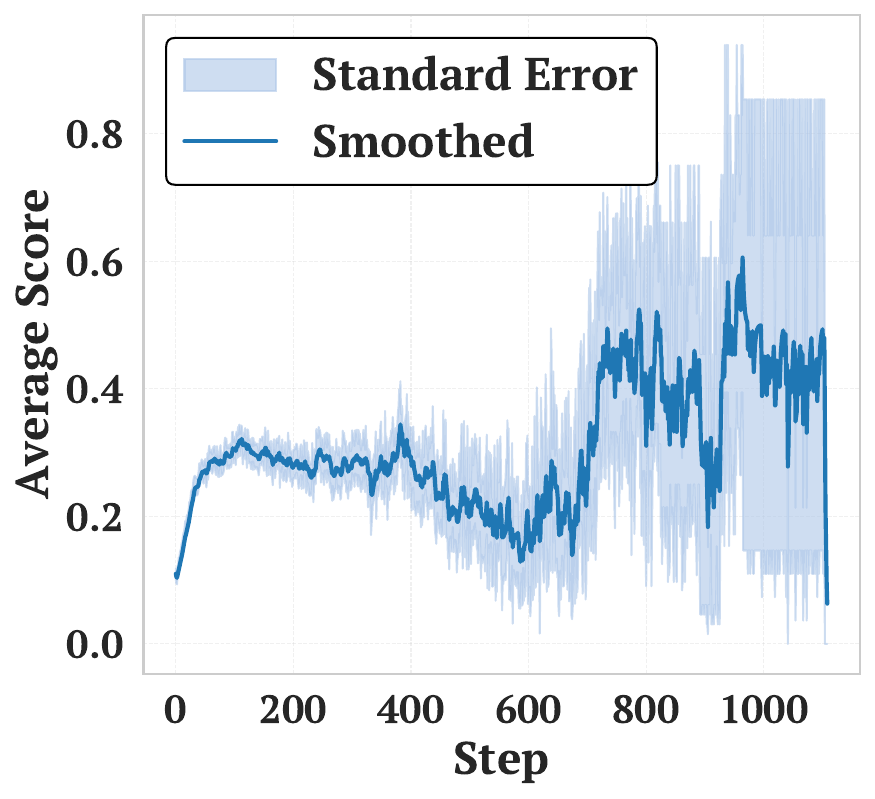}
        \caption{Qwen2.5-3B-Instruct (LIMO)}
    \end{subfigure}

    \begin{subfigure}[b]{0.32\linewidth}
        \centering
        \includegraphics[width=\linewidth]{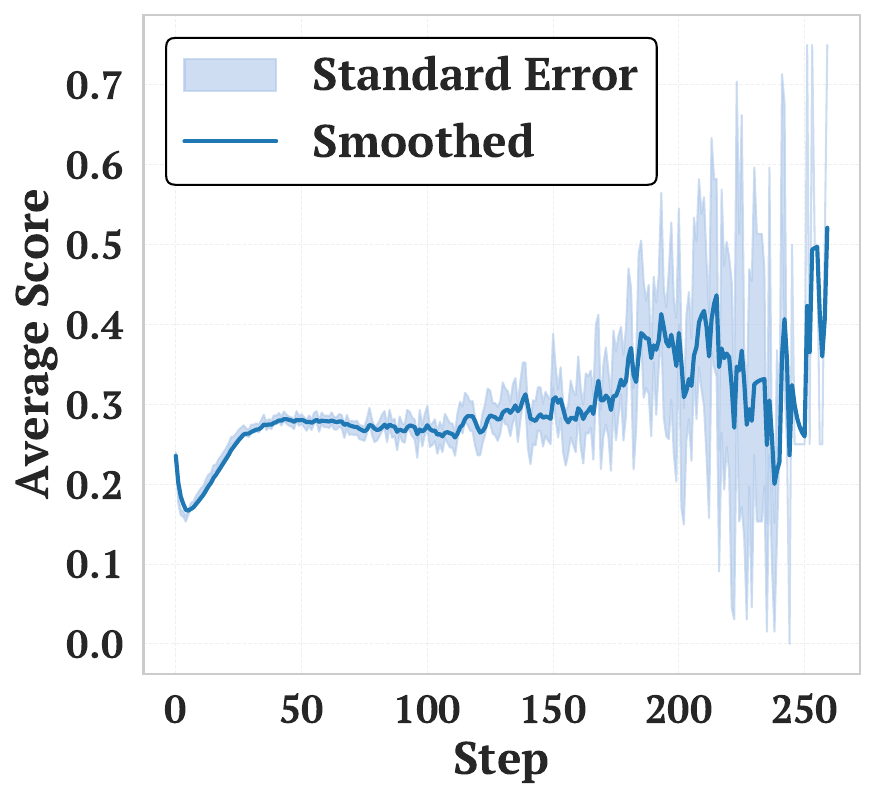}
        \caption{Qwen2.5-0.5B-Instruct (Math-QwQ-32B)}
    \end{subfigure}
    \begin{subfigure}[b]{0.32\linewidth}
        \centering
        \includegraphics[width=\linewidth]{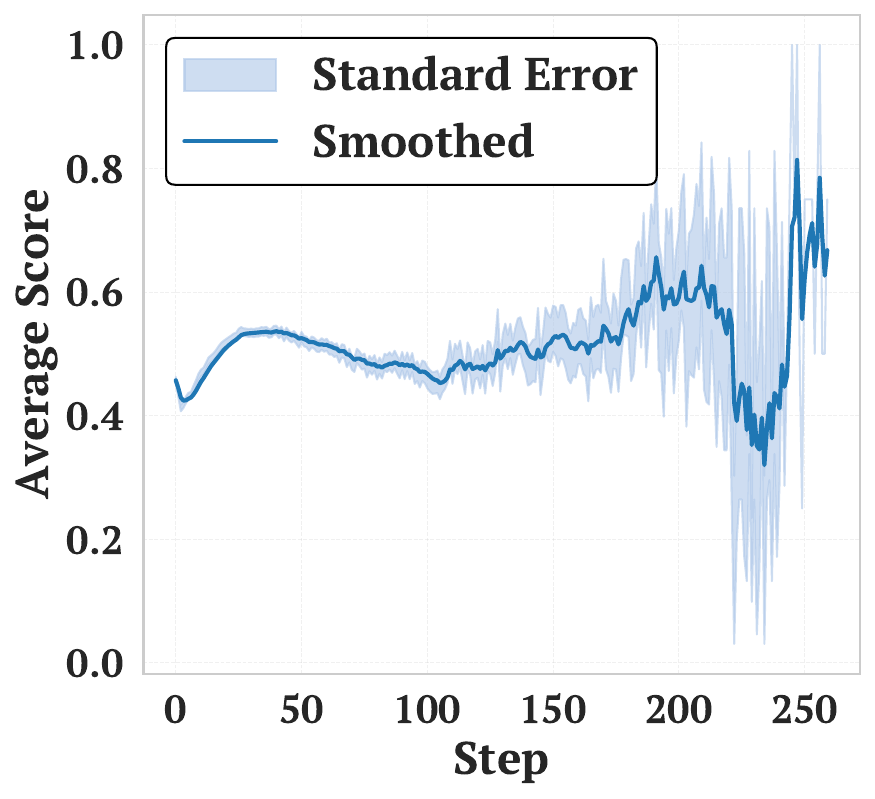}
        \caption{Qwen2.5-1.5B-Instruct (Math-QwQ-32B)}
    \end{subfigure}
    \begin{subfigure}[b]{0.32\linewidth}
        \centering
        \includegraphics[width=\linewidth]{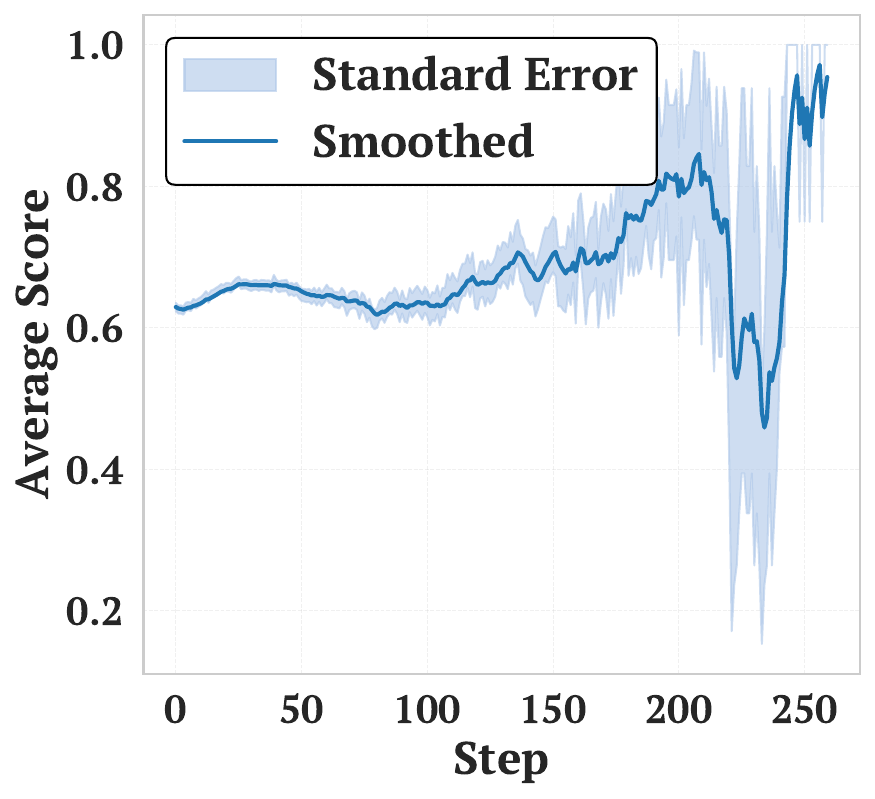}
        \caption{Qwen2.5-3B-Instruct (Math-QwQ-32B)}
    \end{subfigure}
    \label{fig:adaptability_scores}
    \vspace{-1em}
\end{figure}

To empirically validate the \textit{imitation gap} hypothesis introduced in Section~\ref{sec:methodology}, we estimate the step-wise adaptability scores of student models across three parameter scales (0.5B, 1.5B, 3B) on two reasoning datasets (LIMO and Math-QwQ-32B). Each adaptability score quantifies the model’s average probability of reaching the correct final answer when conditioned on imitating a given intermediate step from the expert trajectory. 

As shown in Figure~\ref{fig:adaptability_scores}, these curves reveal a consistent behavioral pattern: early in the reasoning path, student models exhibit increasing adaptability as they benefit from following expert steps. However, beyond a certain point, adaptability scores sharply decline—signaling that the student has encountered steps that exceed its reasoning capacity, leading to degraded rollout completions and a collapse in trajectory success. Notably, toward the final steps—when the expert trajectory approaches the correct answer—the adaptability scores begin to recover, suggesting the model can re-anchor itself when nearing outcome information. 

This non-monotonic pattern reveals the \textbf{imitation gap}—a critical region where student models falter due to misalignment between their capabilities and the expert's step distribution. This misalignment arises from distributional discrepancies, where expert trajectories include reasoning patterns outside the student's abstraction space. Consequently, continued imitation in this zone not only fails to benefit learning but actively impairs performance. This phenomenon underscores our central argument: effective reasoning supervision must be dynamically aligned with model-specific capabilities, as realized in our \methodname{} framework.


\subsection{Impact of Search Restriction on Adaptation Strategies} 
\label{search_restriction}
To evaluate the impact of adaptation strategies without autonomous search, we designed two variants: \textbf{Adaptation-First} and \textbf{Adaptation-Gap}. \textbf{Adaptation-First} halts imitation once a feasible solution state is detected, whereas \textbf{Adaptation-Gap} monitors adaptability scores and terminates imitation when sharp declines occur, as described in Section~\ref{sec:imitation_gap}. Table~\ref{tab:adaptation_results} presents the evaluation results on Math-QwQ-32B for 1.5B and 3B models. Both strategies exhibit performance degradation compared to \textbf{Adaptation-Full}, highlighting the critical role of autonomous search for recovery in complex reasoning paths. Notably, \textbf{Adaptation-Gap} consistently outperforms \textbf{Adaptation-First} across all benchmarks, with significant accuracy gains in the average performance (\textbf{32.8\% vs. 27.6\%} for 1.5B and \textbf{41.1\% vs. 29.0\%} for 3B).
 This improvement stems from its capacity-aware truncation, which effectively filters out low-adaptability segments, preventing error propagation and enhancing stability.


\vspace{-0.5em}
\begin{table}[t]
  \centering
  \small
\caption{
    Accuracy (\%) on the Math-QwQ-32B dataset for 1.5B and 3B models under different adaptation strategies. 
    \textbf{Adaptation-First} performs early stopping at feasible solution states, while \textbf{Adaptation-Gap} selectively truncates imitation paths based on adaptability declines. 
    \textbf{Adaptation-Full} integrates autonomous search, achieving the highest performance across benchmarks. 
    \textbf{Bold} values indicate the best results in each group.
}
  \resizebox{\textwidth}{!}{%
  \begin{tabular}{llcccccccc}
    \toprule
    \thead{Model} & \thead{Method} & \thead{GSM8K} & \thead{MATH} & \thead{Minerva\\Math} & \thead{GaoKao\\2023 En} & \thead{Olympiad\\Bench} & \thead{College\\Math} & \thead{MMLU\\STEM} & \thead{Avg.} \\
    \midrule
    \multicolumn{10}{c}{\textbf{Math-QwQ-32B Dataset}} \\
    \midrule
    \multirow{3}{*}{1.5B}& Adaptation-First                & 41.2 & 30.9 & 11.0 & 33.0 & 10.7 & 24.9 & 41.7 & 27.6 
\\
         & Adaptation-Gap & 60.0 & 37.0 & 13.2 & 34.5 & 13.6 & 29.7 & 41.3 & 32.8 
\\
         & \textbf{Adaptation-Full} & \textbf{72.1} & \textbf{49.9} & \textbf{17.6} & \textbf{43.1} & \textbf{17.9} & \textbf{38.4} & \textbf{44.1} & \textbf{40.4} 
\\
    \midrule     
    \multirow{3}{*}{3B}& Adaptation-First                & 36.7 & 32.2 & 15.4 & 32.7 & 12.9 & 26.1 & 46.8 & 29.0 
\\
         & Adaptation-Gap & 77.9 & 48.6 & 19.5 & 43.4 & 18.4 & 36.0 & 44.0 & 41.1 
\\
         & \textbf{Adaptation-Full} & \textbf{86.6} & \textbf{65.3} & \textbf{26.5} & \textbf{57.9} & \textbf{29.0} & \textbf{44.4} & \textbf{50.6} & \textbf{51.5 }
\\
    \bottomrule
  \end{tabular}
  }
  \vspace{-4mm}
  \label{tab:adaptation_results}
\end{table}

\subsection{Capacity-Aligned Lexical Dynamics Under Adaptation} 
\label{sec:data_shift}
\vspace{-0.5em}

To investigate how \methodname{} reshapes student model behavior at different scales, we analyze keyword frequency changes between static and adapted dataset. Table~\ref{table:freq-change-all} lists the top 20 tokens with the largest shifts in the first sentence of each reasoning step for 0.5B, 1.5B, and 3B models.  

\begin{table}[t]
\centering
\small
\caption{Top 20 Keyword Frequency Changes Across Model Sizes}
\renewcommand{\arraystretch}{1.0}
    \resizebox{\textwidth}{!}{  
    \begin{tabular}{lccc|ccc|ccc}
    \toprule
    \textbf{Keyword} & \multicolumn{3}{c|}{\textbf{0.5B (\%)}} & \multicolumn{3}{c|}{\textbf{1.5B (\%)}} & \multicolumn{3}{c}{\textbf{3B (\%)}} \\
    \cmidrule(lr){2-4} \cmidrule(lr){5-7} \cmidrule(lr){8-10}
    & Static & Adapted & \textbf{$\Delta$} & Static & Adapted & \textbf{$\Delta$} & Static & Adapted & \textbf{$\Delta$} \\
    \midrule
    but           & 2.73 & 2.59 & \cellcolor{red!10}--0.14 & 2.73 & 2.37 & \cellcolor{red!20}--0.36 & 2.73 & 2.27 & \cellcolor{red!30}--0.46 \\
    alternatively & 0.86 & 0.79 & \cellcolor{red!5}--0.07 & 0.86 & 0.72 & \cellcolor{red!10}--0.14 & 0.86 & 0.71 & \cellcolor{red!15}--0.15 \\
    wait          & 2.30 & 2.23 & \cellcolor{red!5}--0.07 & 2.30 & 2.10 & \cellcolor{red!10}--0.20 & 2.30 & 2.00 & \cellcolor{red!15}--0.30 \\
    therefore     & 1.55 & 1.50 & \cellcolor{red!5}--0.05 & 1.55 & 1.43 & \cellcolor{red!10}--0.13 & 1.55 & 1.40 & \cellcolor{red!10}--0.15 \\
    check         & 0.51 & 0.47 & \cellcolor{red!5}--0.04 & 0.51 & 0.40 & \cellcolor{red!10}--0.11 & 0.51 & 0.33 & \cellcolor{red!20}--0.18 \\
    another       & 0.29 & 0.26 & \cellcolor{red!5}--0.03 & 0.29 & 0.20 & \cellcolor{red!10}--0.09 & 0.29 & 0.17 & \cellcolor{red!15}--0.12 \\
    then          & 0.97 & 0.94 & \cellcolor{red!5}--0.03 &  -   &  -   &           -             &  -   &  -   &           -             \\
    pi            & 0.11 & 0.09 & \cellcolor{red!5}--0.02 &  -   &  -   &           -             &  -   &  -   &           -             \\
    perhaps       & 0.55 & 0.53 & \cellcolor{red!5}--0.02 &  -   &  -   &           -             &  -   &  -   &           -             \\
    length        & 0.22 & 0.24 & \cellcolor{green!5}+0.02 &  -   &  -   &           -             &  -   &  -   &          -              \\
    step          & 0.27 & 0.30 & \cellcolor{green!5}+0.02 & 0.27 & 0.37 & \cellcolor{green!10}+0.10 & 0.27 & 0.41 & \cellcolor{green!20}+0.14 \\
    now           & 0.43 & 0.46 & \cellcolor{green!5}+0.03 & 0.43 & 0.49 & \cellcolor{green!10}+0.06 & 0.43 & 0.50 & \cellcolor{green!10}+0.07 \\
    first         & 0.89 & 0.92 & \cellcolor{green!5}+0.03 & 0.89 & 0.96 & \cellcolor{green!10}+0.07 & 0.89 & 0.96 & \cellcolor{green!10}+0.07 \\
    since         & 0.80 & 0.83 & \cellcolor{green!5}+0.03 &  -   &  -   &           -             &  -   &  -   &          -              \\
    have          & 0.64 & 0.67 & \cellcolor{green!5}+0.03 & 0.64 & 0.69 & \cellcolor{green!5}+0.05  &  -   &  -   &        -                \\
    let           & 1.83 & 1.86 & \cellcolor{green!5}+0.04 & 1.83 & 1.88 & \cellcolor{green!5}+0.05  &  -   &  -   &        -                \\
    need          & 0.54 & 0.58 & \cellcolor{green!5}+0.04 & 0.54 & 0.68 & \cellcolor{green!10}+0.14 & 0.54 & 0.69 & \cellcolor{green!10}+0.16 \\
    find          & 0.37 & 0.42 & \cellcolor{green!5}+0.04 & 0.37 & 0.50 & \cellcolor{green!10}+0.13 & 0.37 & 0.54 & \cellcolor{green!15}+0.16 \\
    newline       &  -   &  -   &           -             & 0.00 & 0.05 & \cellcolor{green!5}+0.05  &  -   &  -   &         -               \\
    equation      &  -   &  -   &           -             & 0.76 & 0.81 & \cellcolor{green!5}+0.05  & 0.76 & 0.85 & \cellcolor{green!10}+0.09 \\
    \bottomrule
    \end{tabular}
}
\vspace{-3mm}
\label{table:freq-change-all}  
\end{table}


Adaptation reduces exploratory terms like \textit{but}, \textit{wait}, and \textit{alternatively}, while amplifying goal-oriented expressions such as \textit{step}, \textit{solve}, \textit{find}, and \textit{need}. In the 1.5B model, \textit{but} and \textit{wait} drop by 0.36\% and 0.20\% percentage points, while \textit{find} and \textit{need} rise by 0.13\% and 0.14\% points. This shift reflects a transition from hesitant exploration to decisive, solution-driven reasoning. These changes reduce uncertainty and digression—traits often seen in expert trajectories but burdensome for smaller models. In static supervision, such expressions appear frequently, straining low-capacity models and widening the \textit{Imitation Gap} (Sec.~\ref{sec:mcts_feasibility}), where expert strategies exceed model capabilities.  


\methodname{} bridges this gap by replacing brittle reasoning paths with model-originated decision traces. This adaptation maintains task objectives while restructuring execution to fit the model's capacity, leading to more stable and efficient reasoning.

\section{Related Work}
\vspace{-0.5em}
\paragraph{Chain-of-Thought Reasoning}
Early work on chain-of-thought reasoning (CoT)~\cite{wei2022chain} primarily focused on \textit{short CoT}, where models generate concise reasoning paths to solve problems. Recent advances~\cite{chen2025reasoningerasurveylong} have shifted towards \textit{long CoT prompting}, encouraging more elaborate reasoning chains that enable systematic exploration of multiple paths (\textit{branching}) and backtracking when errors are detected. While techniques like knowledge distillation(~\cite{hinton2015distill,luo2025longcotdistill}) and reinforcement learning~\cite{hou2025thinkprune} have been used to equip large language models (LLMs) with long CoT capabilities, these efforts remain largely confined to models with substantial parameter sizes. In contrast, our work specifically addresses the unique challenges associated with training smaller-scale models for complex reasoning tasks.

\vspace{-0.5em}
\paragraph{Data-Efficient Reasoning Elicitation}
A related line of work investigates how minimal supervision can elicit latent reasoning abilities in pretrained models~\cite{ye2025limoreasoning,muennighoff2025s1simpletesttimescaling}. These methods rely on a few carefully designed \emph{cognitive templates}, to guide reasoning, but often assume that models possess the necessary prior knowledge. This assumption makes the templates brittle when cognitive demands exceed model capacity. To address this limitation, we propose a feasibility-aware adaptation framework that dynamically adjusts supervision to model ability, enabling robust reasoning across diverse capacity profiles.

\section{Conclusion}
\vspace{-0.5em}
\label{sec:conclusion}
We propose \methodnamefull{} (\methodname{}), a data adaptation framework designed to improve reasoning elicitation for small language models. By introducing adaptability-based selective imitation and outcome-consistent exploration, our method aims to better align expert demonstrations with model capabilities. Experimental results across several benchmarks show that \methodname{} can improve reasoning performance compared to static fine-tuning. We hope this work provides a step toward more flexible and model-aware data alignment strategies for reasoning tasks.

\vspace{-0.5em}
\section{Limitations and future work}
\label{sec:limitations_future}
\vspace{-0.5em}
Our framework is effective for structured reasoning tasks with verifiable outcomes. However, its extension to open-ended tasks with inherent output uncertainty remains limited, suggesting the need for refined supervision mechanisms and evaluation metrics to ensure outcome consistency.


\bibliographystyle{plainnat}
\bibliography{ref}

\newpage
\section*{Appendix}
\setcounter{subsection}{0}
\renewcommand{\thesubsection}{\Alph{subsection}}

\subsection{Experimental Details}
\label{sec:exp_details}
\subsubsection{Experiment Prompts}
In our simulation experiments, we employed a structured prompting approach to guide the language model through multi-step reasoning tasks. The primary simulation prompt used in our study is defined as follows:

\begin{tcolorbox}[colback=gray!5!white, colframe=gray!75!black,
  title=Simulation Prompt, boxrule=0.5pt, arc=3pt, left=2mm, right=2mm, top=1mm, bottom=1mm]
\textbf{Question:} [Question] \\
\textbf{Rationale so far:} $s_1, s_2, \ldots, s_{t_{\text{n}}}$ \\
Continue reasoning step by step, and put your final answer within \textbackslash boxed\{\}.
\end{tcolorbox}

This prompt encourages the model to decompose the problem into intermediate steps and to clearly indicate the final answer using LaTeX-style boxed notation. This formatting ensures consistency across outputs and facilitates automated evaluation of results.

In addition to standard simulation prompting, we introduce a dedicated exploration prompt tailored for the adaptive trajectory rollout described in Section~\ref{sec:adaptive_exploration}. This prompt is activated once a low-adaptability segment is detected and aims to continue reasoning beyond the imitation gap. It conditions the model on the prefix of high-adaptability reasoning steps and allows for autonomous continuation constrained only by outcome correctness:

\begin{tcolorbox}[colback=gray!5!white, colframe=gray!75!black,
  title=Exploration Prompt, boxrule=0.5pt, arc=3pt, left=2mm, right=2mm, top=1mm, bottom=1mm]
\textbf{Question:} [Question] \\
\textbf{Rationale before the gap:} $s_1, s_2, \ldots, s_{t_{\text{gap}} - 1}$ \\
Continue reasoning step by step, and put your final answer within \textbackslash boxed\{\}.
\end{tcolorbox}

This exploration prompt encourages the model to develop its own reasoning path from the last trustworthy segment, fostering flexible generalization while maintaining semantic alignment with the expert outcome.

\subsubsection{Parameter Configuration for Simulation}

The simulation procedure in Algorithm~\ref{sec:mcts_feasibility} adopts stochastic decoding to explore alternative reasoning paths beyond expert demonstrations. We sample $N = 4$ candidate continuations per step, corresponding to the adaptation simulation count $N_{\text{sim}}$.

Each trajectory is generated with a maximum length of \texttt{MAX\_NEW\_TOKENS = 4000}. To promote determinism while retaining minimal stochasticity, we set the sampling temperature to \texttt{TEMPERATURE = 0.1}. Decoding is performed in batches of \texttt{BATCH\_SIZE = 32} to enable efficient parallel inference under hardware constraints.

These settings ensure stable simulation rollouts with low-variance outputs, suitable for evaluating adaptability under controlled decoding conditions.

\subsubsection{Parameter Configuration for Adaptive Path Exploration}

To support the adaptive rollout mechanism described in Section~\ref{sec:adaptive_exploration}, we configured the \textsc{Explore} phase with carefully selected hyperparameters to balance computational efficiency and response diversity. The sampling procedure was executed with a candidate beam size of \texttt{NUM\_SAMPLES = 8}, meaning that at each decision step, eight reasoning continuations were generated for evaluation based on the adaptability score.

We set the maximum generation length to \texttt{MAX\_NEW\_TOKENS = 2000} to allow sufficient space for multi-step reasoning without premature truncation. A temperature of \texttt{TEMPERATURE = 0.7} was employed to introduce moderate randomness in token sampling, facilitating the exploration of alternative reasoning paths while retaining coherence.

Batch inference was performed with a \texttt{BATCH\_SIZE = 64} to utilize GPU resources efficiently during large-scale rollouts. The underlying language model was run using half-precision arithmetic (\texttt{DTYPE = float16}), which reduced memory footprint and improved throughput without compromising output quality.

Additionally, the maximum number of concurrent sequences handled by the inference engine (VLLM) was set to \texttt{MAX\_NUM\_SEQS = 512}, enabling high-throughput parallel generation during exploration. These settings ensured scalable, stable, and semantically diverse adaptation rollouts that align with the outcome consistency constraint described in Equation~\eqref{eq:outcome}.

\subsection{Impact of Search Path Quality on Model Performance}

\begin{table}[t]
  \centering
    \caption{
        Comparison of accuracy (\%) on the Math-QwQ-32B dataset for 0.5B and 1.5B models under different adaptation strategies.The table contrasts the performance between \textit{Adaptation-Raw} (without removing repeated paths) and \textit{Adaptation-Cleaned} (with repeated paths removed). \textbf{Bold} values indicate the best results.
    }
  \resizebox{\textwidth}{!}{%
  \begin{tabular}{llcccccccc}
    \toprule
    \thead{Model} & \thead{Method} & \thead{GSM8K} & \thead{MATH} & \thead{Minerva\\Math} & \thead{GaoKao\\2023 En} & \thead{Olympiad\\Bench} & \thead{College\\Math} & \thead{MMLU\\STEM} & \thead{Avg.} \\
    \midrule
    \multicolumn{10}{c}{\textbf{Math-QwQ-32B Dataset}} \\
    \midrule
    \multirow{3}{*}{0.5B}& Adaptation-Raw         & 47.5 & 34.3 & 7.0 & 29.1 & 9.3 & 26.8 & 28.2 & 26.0 
\\
                         & Adaptation-Cleaned     & \textbf{49.6} & \textbf{34.0} & \textbf{6.2} & \textbf{30.9} & \textbf{8.4} & \textbf{27.5} & \textbf{37.5} & \textbf{27.7} 
\\
    \midrule     
    \multirow{3}{*}{1.5B}& Adaptation-Raw         & 70.8 & 48.9 & 15.8 & 44.2 & 18.8 & 38.3 & 44.6 & 40.2 
\\
                         & Adaptation-Cleaned     & \textbf{74.2} & \textbf{55.1} & \textbf{17.6} & \textbf{48.6} & \textbf{19.6 }& \textbf{39.4} & \textbf{57.7} & \textbf{44.6 }
\\
    \bottomrule
  \end{tabular}
  }
  \vspace{-4mm}
  \label{tab:rawvsclean}
\end{table}

\begin{figure}[htbp]
  \centering
  \includegraphics[width=0.7\textwidth]{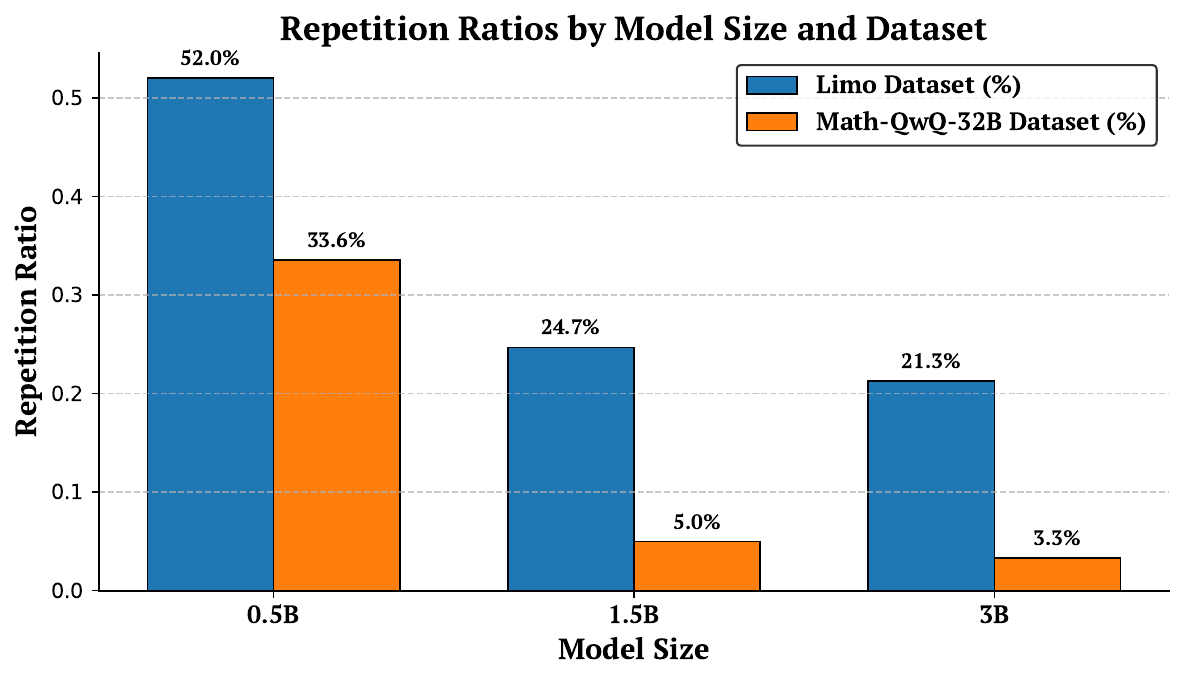}
  \caption{Repetition ratios(\%) in search paths across different model sizes and datasets. Smaller models tend to have higher repetition ratios, particularly on the Limo dataset.}
  \label{fig:repetition_ratio}
\end{figure}



To investigate the impact of search quality on model performance, we conducted a comparative experiment (see Table~\ref{tab:rawvsclean}). After completing the adaptation path search, we removed paths exhibiting severe repetition phenomena. As illustrated in Figure~\ref{fig:repetition_ratio}, the proportion of repeated paths during exploration decreases progressively with increasing model parameter size, indicating that improvements in the model’s generative capability and contextual memory effectively reduce repetition. 

We refer to the results after removing such repeated paths as \textit{Adaptation-Cleaned} and systematically evaluated these against the complete search results without removing repeated paths, denoted as \textit{Adaptation-Raw}. Experimental results demonstrate that filtering out repeated paths leads to significant performance gains, further highlighting the critical role of search path quality in overall model performance.

\subsection{Comparative Analysis of Truncation Methods under Search Constraints}

\begin{table}[t]
  \centering
  \small
  \caption{
    Comparison of truncation positions between \textbf{Adaptation-First} and \textbf{Adaptation-Gap} methods across datasets and model sizes. The relative localization difference represents the absolute difference between the relative truncation positions of these two methods. Higher differences are highlighted with deeper red.
  }
  \resizebox{\textwidth}{!}{%
  \begin{tabular}{lccccc}
    \toprule
    \textbf{Dataset} & \textbf{Model Size} & \textbf{First Position} & \textbf{Gap Position} & \textbf{Relative Localization Difference} \\
    \midrule
    \multirow{3}{*}{Limo Dataset} 
      & 0.5B & 0.7901 & 0.7707 & \coloredcell{0.0194} \\
      & 1.5B & 0.6785 & 0.6985 & \coloredcell{0.0200} \\
      & 3B   & 0.5718 & 0.5983 & \coloredcell{0.0265} \\
    \midrule
    \multirow{3}{*}{Math-QwQ-32B Dataset} 
      & 0.5B & 0.5055 & 0.5244 & \coloredcell{0.0189} \\
      & 1.5B & 0.2113 & 0.3664 & \coloredcell{0.1551} \\
      & 3B   & 0.0840 & 0.3239 & \coloredcell{0.2399} \\
    \bottomrule
  \end{tabular}
  }
  \vspace{-4mm}
  \label{tab:truncation_comparison}
\end{table}

In our previous section (see Section~\ref{search_restriction}), we investigate two truncation methods under different search constraints. Specifically, we designed two variants: \textbf{Adaptation-First} and \textbf{Adaptation-Gap}. The \textbf{Adaptation-First} method halts imitation once a feasible solution state is detected, whereas \textbf{Adaptation-Gap} monitors adaptability scores and terminates imitation when sharp declines occur, as detailed in Section~\ref{sec:imitation_gap}.

We compare the truncation positions of the two methods across different datasets and model sizes. Our analysis indicates that on more challenging datasets, or when the model capacity is limited (e.g., results on the 0.5B models for both datasets), the truncation points identified by \textbf{Adaptation-First} and \textbf{Adaptation-Gap} are largely consistent. This can be attributed to the complexity of the reasoning cognitive templates in these datasets relative to the model’s capabilities: once the model identifies a path leading to a feasible solution, continued imitation often ventures into regions that are difficult to adapt to, typically accompanied by a sharp decline in adaptability scores. Consequently, the truncation positions under both \textbf{Adaptation-First} and \textbf{Adaptation-Gap} modes are generally aligned.

Conversely, on the Math-Qwen dataset, notable differences in truncation positions emerge. Many models, after reaching the step at which the final answer can be searched, continue to utilize subsequent adaptable path segments. Thus, the \textbf{Adaptation-Gap} method is able to detect and leverage a greater number of these usable step fragments, resulting in more substantial performance improvements, as reported in Table~\ref{tab:adaptation_results}.



\end{document}